\documentclass[runningheads]{llncs}
\usepackage[T1]{fontenc}
\usepackage{graphicx}
\usepackage{amssymb}
\usepackage[absolute,overlay]{textpos}
\usepackage{longtable}\usepackage{longtable}
\usepackage{array}
\newcolumntype{C}{>{\centering\arraybackslash}m{2.3cm}}
\newcolumntype{D}{>{\centering\arraybackslash}m{1.15cm}}
\newcolumntype{E}{>{\centering\arraybackslash}m{1.25cm}}
\newcolumntype{F}{>{\centering\arraybackslash}m{0.85cm}}
\newcolumntype{G}{>{\centering\arraybackslash}m{2.25cm}}
\newcolumntype{K}{>{\raggedright\arraybackslash}m{3.75cm}}
\usepackage[export]{adjustbox}
\usepackage{multirow}
\usepackage{amsmath}
\usepackage{xcolor}

\begin{document}
\title{Spatial Graph Convolution Neural Networks for Water Distribution Systems}
%
%
\author{Inaam Ashraf\orcidID{0000-0001-9841-3628} \and
Luca Hermes\orcidID{0000-0002-7568-7981} \and
Andr\'e Artelt\orcidID{0000-0002-2426-3126} \and
Barbara Hammer\orcidID{0000-0002-0935-5591}}
\authorrunning{I. Ashraf et al.}
%
\institute{Center for Cognitive Interaction Technology, Bielefeld University, Bielefeld, Germany\\
\email{\{mashraf, lhermes, aartelt, bhammer\}@techfak.uni-bielefeld.de}}
\maketitle              
\begin{abstract}
\vspace{-.5cm}
We investigate the task of missing value estimation in graphs as given by water distribution systems (WDS) based on sparse 
signals as a representative machine learning challenge in the domain of critical infra\-structure. The underlying graphs have a comparably low node degree and high diameter, while information in the graph is globally relevant, hence graph neural networks face the challenge of long term dependencies. We propose a specific architecture  based on message passing which displays excellent results for a number of benchmark tasks in the WDS domain. Further, we  investigate a multi-hop variation, which requires considerably less resources and opens an avenue towards big WDS graphs.

\end{abstract}

\section{Introduction}
Transportation systems, energy grids, and water distribution systems (WDS) constitute parts of our critical infrastructure that are vital to our society and subject to special protective measures and regulations. As they are under increasing strain in the face of limited resources 
and as they are vulnerable to attacks, their efficient management and continuous monitoring is of great importance.
As an example, the average amount of non-revenue water amounts to 25\% in the EU \cite{watereau}, making the detection of leaks in WDS an important task.
Advances in sensor technology and increasing digitalisation hold the potential for intelligent monitoring and adaptive control using AI technologies \cite{pmlr-v176-eichenberger22a,smartcities4020029,doi:10.1061/(ASCE)PS.1949-1204.0000646}. 
In addition to more classical AI approaches, deep learning technologies are increasingly being used to solve  learning tasks in the context of critical infrastructures
\cite{doi:10.1061/(ASCE)IS.1943-555X.0000477}.

A common feature of WDS, energy networks and transport networks is that the data has a temporal and spatial character: Data is generated in real time according to an underlying graph, given by the power grid, the pipe network and the transport routes, respectively. Measurements  are available for some nodes that correspond to local sensors, e.g.\ pressure sensors or smart meters. Based on this partial information, the task is to derive corresponding quantities at every node of the graph, to identify the system state  or to derive optimal planning and control strategies.
In this paper, we target the learning challenges of the first feature, inferring relevant quantities at each location of the graph based on few measurements.
While classical deep learning models such as convolutional networks or recurrent models can reliably handle Euclidean data, graphs constitute non-Euclidean data that require techniques from geometric deep learning.
Based on initial approaches dating back more than a decade
 \cite{scarselli2009,DBLP:journals/neco/HammerMS05}, 
 a variety of graph neural networks (GNNs) have recently been proposed that are able to directly process information such as present in critical infrastructure \cite{Bruna2014SpectralNA,defferrard2016convolutional,gao2018large,kipf2017semi,velickovic2018graph}. 
 First applications demonstrate the suitability of GNNs for the latter
\cite{https://doi.org/10.48550/arxiv.2202.08065,pmlr-v176-eichenberger22a,defferrard2016convolutional}. 

Graphs from the domain of WDS or smart grids display specific characteristics (s. Fig.~\ref{graph_L-Town_normal}):
as they are located in the plane, the node degree is small and the network diameter is large. These characteristics display a challenge for GNNs, as the problem of long-term dependencies and over-smoothing occurs  \cite{xu2018powerful,DBLP:journals/corr/abs-2003-04078}.
In this contribution, we design a GNN architecture capable of dealing with these specific graph structures: 
We show that our spatial GNN is able to effectively integrate long-range node dependencies
and we demonstrate the impact of a suitable transfer function and residual connections. As the required resources quickly become infeasible for big graphs, we also investigate the comparability of a sparse multi-hop alternative. All methods are evaluated for  pressure prediction  in WDS for a variety of benchmark networks, displaying promising results.

\section{Related Work}

The task of pressure estimation at all nodes in a WDS from pressure values available at a few nodes has recently been dealt with \cite{hajgato2021pressure}. The authors employed spectral graph convolutional neural networks (GCNs) and performed extensive experiments to demonstrate their approach. However, their methodology does not fully benefit from the available structural information of the graph; we provide further details on this in Sec. \ref{sec:experiments}. We propose a spatial GCN based methodology that effectively utilizes the graph structure by using both node and edge features and thus produces significantly better results (s. Sec. \ref{Results}).  

A related task of state (pressure, flow) estimation in WDS based on demand patterns and sparse pressure information has been addressed \cite{xing2022stateestimation}. The authors used hydraulics in the optimization objective since the task was to model the complex hydraulics used by the popular EPANET simulator \cite{rossman2020epanet} using GNNs. They present promising results only on relatively small WDS, the ability to scale to larger WDS is yet to be investigated. While their model solves the task of state estimation in WDS, their approach requires demand patterns from every consumer also during inference. In contrast, our proposed model relies on pressure values computed by the EPANET solver (based on demand patterns) only during the training process. During evaluation, our model estimates pressures solely based on sparse pressure values obtained from a few sensors. Further, it successfully estimates pressures even in case of noisy demands (s. Sec. \ref{Results}).

GNNs were first introduced in the work \cite{scarselli2009} as an extension of recursive neural networks for tree structures \cite{diss}. Since then, a number of GCN algorithms have been developed, which can be classified in to spectral-based and spatial-based. The approach \cite{Bruna2014SpectralNA} introduced spectral GCNs based on spectral graph theory, which was followed by further work \cite{kipf2017semi,defferrard2016convolutional,henaff2015deep,levie2018cayleynets,li2018adaptive}. The counterpart are spatial GCNs which apply a local approximation of the spectral graph kernel  \cite{hamilton2017inductive,monti2017geometric,gao2018large,niepert2016learning,xu2018powerful,velickovic2018graph}. These are also referred to as message passing neural networks.

Unlike convolutional neural networks (CNNs), spatial GCNs suffer from issues like vanishing gradient, over-smoothing and over-fitting, when used to build deeper models. Generalized aggregation functions, residual connections and normalization layers can address these issues and improve performance on diverse GCN tasks and large scale graph datasets \cite{li2020GENConv} . 

To enable high-level embeddings in feed-forward neural networks, self normalizing neural networks (SNNs) were introduced \cite{Gunter2017SNNs}
based on a special activation function called scaled exponential unit (SeLU). 
We combine residual connections \cite{li2020GENConv} with SNNs since residual connections help solve the over-smoothing problem when we use multiple GCN layers, whereas self-normalizing property of SeLU enables the required information propagation in case of sparse features. 

\section{Methodology}

\begin{figure}[t]
    \centering
    \includegraphics[width=\textwidth, keepaspectratio]{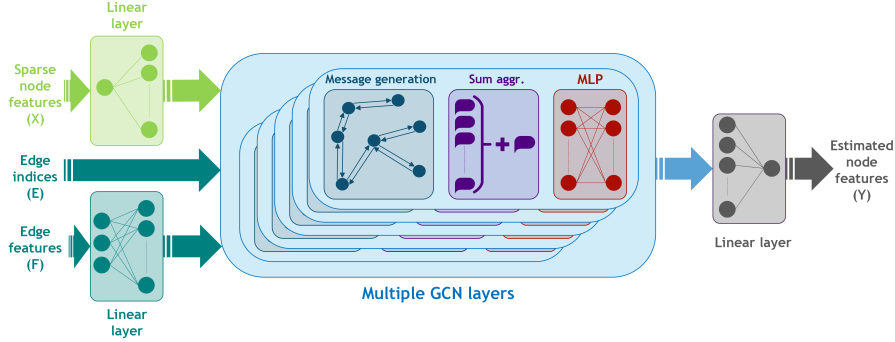}
    \caption{Model architecture employing multiple GCN layers. Each GCN layer consists of message generation, sum aggregation and a final MLP.} \label{model}
    \vspace{-.5cm}
\end{figure}

The main contribution of our work is a spatial GCN  capable of efficiently dealing with the specific graph characteristics as present in WDS. We address the estimation of missing node features based on sparse measurements. As we detail below, we employ multiple spatial GCN layers without suffering from typical problems of vanishing gradient, over-smoothing and over-fitting. For this purpose, we combine residual connections (\cite{li2020GENConv}) with SeLU activation function (\cite{Gunter2017SNNs}). To decrease model size, we leverage GCN layers with multiple hops realizing message passing between more distant neighbors comparable to \cite{li_diffusion_2017}. 
Our model employs spatial GCNs using both node and edge features. 
The complete architecture is depicted in Fig.~\ref{model}. 
Formally, a graph is represented as $G(V,X,E,F)$, where:
\begin{itemize}
    \item $V=\{v_1,v_2,\dots,v_N\}$ is the set of nodes,
    \item $E=\{ e_{vu} \; | \; \forall \, v \in V; u \in \mathcal{N}(v) \}$ is the set of edges, 
    \item $X \in R^{N \times D}$ is the set of node features, where $N = |V|$ and $D$ is the number of node features
    \item $F \in R^{M \times K}$ is the set of edge features, where $M = |E|$ and $K$ is the number of edge features
\end{itemize}
Node   and edge features are embedded 
by fully connected linear layers $\alpha$ and $\beta$:
\begin{equation}
\begin{aligned}[t]
    h_v^{1} &= \alpha (x_v) \\
    h_{e_{vu}}^{1} &= \beta (f_{e_{vu}})
\end{aligned}\quad\quad
\begin{aligned}[t]
    &v \in V, x_v \in X\\ 
    &e_{vu} \in E, f_{e_{vu}} \in F
\end{aligned}
\end{equation}

We denote intermediate model activations as $h_v$ for nodes and $h_{e_{vu}}$ for edges.
Multiple GCN layers convolve the information from the neighboring nodes for estimation of node features. Each GCN layer employs the three-step process of message generation, message aggregation and node feature update. 
In the $l^{th}$ layer, the edge features are updated by
\begin{equation}\label{Edge Update}
    \hat{h}_{e_{vu}}^{(l)} = h_{e_{vu}}^{(l)} + | h_u^{(l)} - h_v^{(l)} |.
\end{equation}

Adding the absolute difference between the current and neighbor nodes features empirically improves the learning. Then, messages are generated as follows:

\begin{equation}
    m_{e_{vu}}^{(l)} = \mathrm{SeLU}\left(h_u^{(l)} \parallel \hat{h}_{e_{vu}}^{(l)}\right) \quad u \in \mathcal{N}(v),
\end{equation}

\noindent where $\cdot \parallel \cdot$ denotes vector concatenation.
After concatenation, we employ the SeLU activation function (\cite{Gunter2017SNNs}) to all messages, which is given by:
\begin{equation}
  \mathrm{SeLU}(x) = \lambda \left\{\begin{array}{rcl} x & if & x > 0 \\ \alpha x - \alpha & if & x <= 0 \end{array}\right.  
\end{equation}
where $\lambda$ and $\alpha$ are hyperparameters as in \cite{Gunter2017SNNs}. SeLU's self-normalizing nature  greatly improves learning in the light of highly sparse values at the beginning of the training process. 
All messages from the neighbor nodes 
are sum-aggregated:
\begin{equation}
    m_v^{(l)} = \sum_{u \in \mathcal{N}(v)} m_{e_{vu}}^{(l)}
\end{equation}
Similar to \cite{li2020GENConv}, we add residual connections to the aggregated messages and pass these through a Multi-Layer Perceptron (MLP):
\begin{equation}
    h_v^{(l+1)} = \mathrm{MLP}\left(h_v^{(l)} + m_v^{(l)}\right)
\end{equation}
The overall message construction, aggregation and update is \cite{li2020GENConv}:
\begin{equation}\label{GENConv Aggregation Objective}
    h_v^{(l+1)} = \mathrm{MLP}\left(h_v^{(l)} + \sum_{u\in\mathcal{N}(v)} \text{SeLU}\left(h_u^{(l)} \parallel \hat{h}_{e_{vu}}^{(l)}\right) \right)
\end{equation}
After employing multiple GCN layers, the resultant node embeddings are fed to a final fully-connected linear layer to estimate all node features. 
\begin{equation}
    \hat{y}_v = \gamma (h_v^{L})\quad\quad v \in V, \hat{y}_v \in \hat{Y}
\end{equation}
where $\hat{Y}$ is the estimated node features, $L$ is the last GCN layer and $\gamma$ is modeled by the linear layer. We use the L1 loss 
as objective function:
\begin{equation}
    \mathcal{L} (y, \hat{y}) = \frac{1}{S\cdot N} \sum_{i=1}^{S} \sum_{v=1}^{N} | y_{iv} - \hat{y}_{iv} | 
\end{equation}
with $Y$ as ground truth, $N$ as  number of nodes and $S$ as number of samples in a mini-batch.

\begin{figure}[t]
    \includegraphics[width=\textwidth, keepaspectratio]{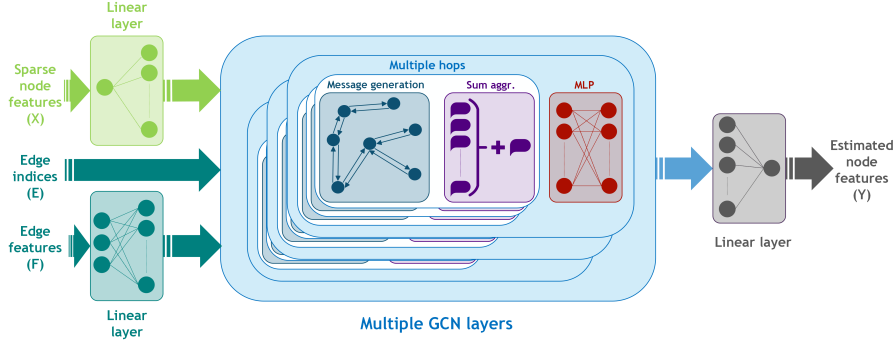}
    \caption{Model architecture employing multiple multi-hop GCN layers.} \label{model_multi_hop}
    \vspace{-.5cm}
\end{figure}

\paragraph{Multi-hop Variation}

Given the sparsity and size of a graph, our methodology requires a comparably large number of  GCN layers  proportional to the size of graph. This reduces scalability to larger graphs. To reduce the number of parameters, we propose GCN layers with multiple hops as shown in Fig.~\ref{model_multi_hop}. Specifically, message generation and aggregation are repeated in each GCN layer before passing it to the MLP:
\begin{equation}
    h_v^{(l)(p+1)} = h_v^{(l)(p)} + \sum_{u \in \mathcal{N}(v)} \mathrm{SeLU}\left(h_u^{(l)(p)} \parallel \hat{h}_{e_{vu}}^{(l)(p)}\right), \quad\quad p \in P
\end{equation}
with $P$ as number of hops.  The embedding for the next layer is: 
\begin{equation}\label{GENConv Multi-hop Aggregation Objective}
    h_v^{(l+1)} = \mathrm{MLP}\left(h_v^{(l)(P)} + \sum_{u \in \mathcal{N}(v)} \mathrm{SeLU}\left(h_u^{(l)(P)} \parallel \hat{h}_{e_{vu}}^{(l)(P)}\right)\right)
\end{equation}
This enables the model to gather information from neighbors that are multiple hops away, requiring fewer GCN layers.

\section{Experiments}\label{sec:experiments}

\begin{figure}[t]
    \includegraphics[width=\textwidth, keepaspectratio]{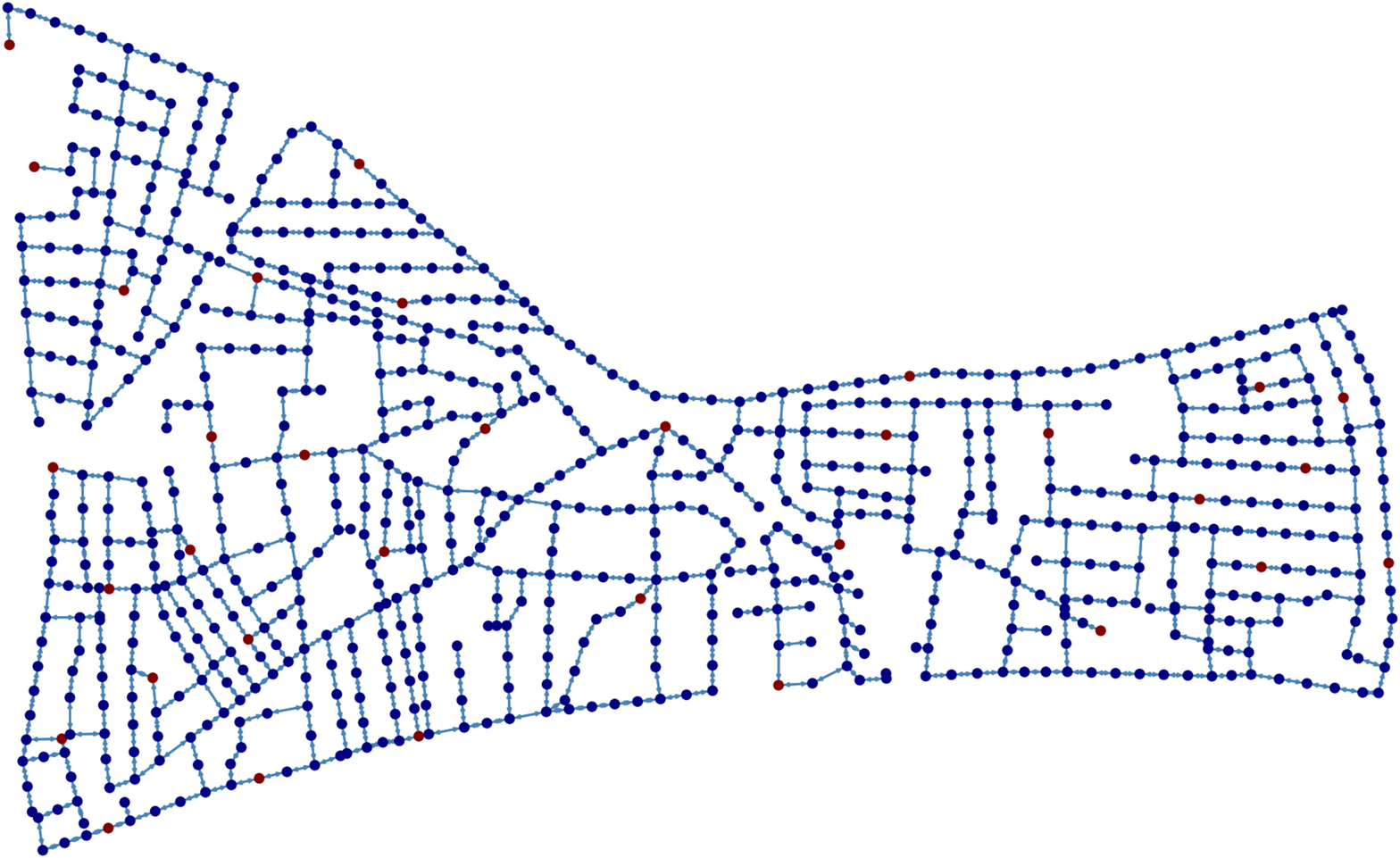}
    \caption{L-Town Water Distribution System (\cite{vrachimis2020battledim}) -- nodes in \textcolor{red}{red}  have sensors.} \label{graph_L-Town_normal}
    \vspace{-.25cm}
\end{figure}

The methodology can be applied to missing node feature estimation on any graph. Here, we investigate WDS, which are modelled as graphs by representing junctions as nodes and pipes between junctions as edges.
WDS are especially challenging because pressure sensors are installed at only few nodes due to constraints (size of the system, cost, availability, practicality) \cite{Klise2013sensor}, resulting in graphs with sparse feature information. Additionally, the node degree in WDS is usually low (s. Tab. \ref{wdn_table}). 
These properties can be observed in the popular L-Town WDS \cite{vrachimis2020battledim} shown in Fig. \ref{graph_L-Town_normal}.
Such characteristics require GNNs to model long-range dependencies between nodes to properly integrate the available information.

\begin{table}[b]
    \vspace{-0.5cm}
    \caption{Major attributes of WDS.}\label{wdn_table}
    \centering
    \resizebox{\textwidth}{!}{
    \begin{tabular}{|K|C|C|C|C|}
    \hline
    WDS & Anytown & C-Town & L-Town & Richmond \\
    \hline
    Number of junctions & 22 & 388 & 785 & 865 \\
    Number of pipes & 41 & 429 & 909 & 79 \\
    Diameter & 5 & 66 & 79 & 234 \\
    Degree (min, mean, max) & (1, 3.60, 7) & (1, 2.24, 4) & (1, 2.32, 5) & (1, 2.19, 4)  \\
    \hline
    \end{tabular}
    }
    \vspace{-0.5cm}
\end{table}

To the best of our knowledge, the task of node feature estimation in WDS using GNNs based on sparse features has only been dealt by \cite{hajgato2021pressure}. These researchers compared their model to a couple of non-GNN based baselines: The first baseline refers to the mean of known node features as value for unknown node features, the second baseline uses interpolated regularization \cite{belkin2004}. The work \cite{hajgato2021pressure} demonstrates that the GNN model significantly outperforms both baselines. Therefore, in our experiments, we compare our approach only to the GNN model, ChebNet, of \cite{hajgato2021pressure}.
We run two experiments on simulated data. First, we compare our approach to \cite{hajgato2021pressure} on three WDS datasets Anytown, C-Town, and Richmond. Second, we conduct an in-depth evaluation on L-Town with extensive hyperparameter tuning.

\vspace{-.25cm}
\subsection{Datasets}

We use a total of four WDS datasets for our experiments: Anytown, C-Town, L-Town and Richmond \footnote{\href{https://engineering.exeter.ac.uk/research/cws/resources/benchmarks/\#a8}{https://engineering.exeter.ac.uk/research/cws/resources/benchmarks/\#a8}} \footnote{\href{https://www.batadal.net/data.html}{https://www.batadal.net/data.html}} (\cite{vrachimis2020battledim}). Major attributes of the WDS are listed in Table \ref{wdn_table}. We use the dataset generation methodology of \cite{hajgato2021pressure} for three of the WDS (Anytown, C-Town, Richmond) and record 1000 consecutive time steps for each of the three networks. For each network, we use three different sparsity levels i.e. sensor ratios of 0.05, 0.1 and 0.2. We do not evaluate on sparsity levels of 0.4 and 0.8 as done in \cite{hajgato2021pressure}, which are more easy.
We sample 5 different random sensor configurations for each sparsity level and each WDS instead of 20.

\begin{table}[t]
    \caption{Model Hyperparameters and Parameters.}\label{hyperparameters}
    \centering
    \resizebox{\textwidth}{!}{
    \begin{tabular}{|p{1.5cm}|p{3cm}|C|C|C|D|D|}
    \hline
    Model &  & Anytown & C-Town & Richmond & \multicolumn{2}{|c|}{L-Town} \\
    \hline
    \multirow{4}{*}{ChebNet} & No. of layers & 4 & 4 & 4 & \multicolumn{2}{|c|}{4} \\
     & Degrees ($K_i$) & [39, 43, 45, 1] & [200, 200, 20, 1] & [240, 120, 20, 1] & \multicolumn{2}{|c|}{[240, 120, 20, 1]}\\
     & No. of filters ($F_i$) & [14, 20, 27, 1] & [60, 60, 30, 1] & [120, 60, 30, 1] & \multicolumn{2}{|c|}{[120, 60, 30, 1]}\\
     & Parameters (million) & 0.038 & 0.780 & 0.958 & \multicolumn{2}{|c|}{0.929}\\
    \hline
    \multirow{5}{*}{m-GCN} & No. of GCN layers & 5 & 33 & 60 & 45 & 10 \\
     & No. of hops & 1 & 2 & 3 & 1 & 5 \\
     & No. of MLP layers & 2 & 2 & 2 & 2 & 2 \\
     & Latent dimension & 32 & 32 & 48 & 96 & 96 \\
     & Parameters (million) & 0.031 & 0.203 & 0.830 & 2.488 & 0.553 \\
    \hline
    \end{tabular}
    }
    \vspace{-.5cm}
\end{table}

For the  popular L-Town network, we use only a single configuration of sensors as designed by \cite{vrachimis2020battledim}, which gives a sensor ratio of 0.0422. We use two different sets of simulation settings; one with 
smooth toy demands and the other close to actual noisy demand patterns. The simulations are carried out using EPANET \cite{rossman2020epanet} provided by Python package \textit{wntr} (\cite{klise2018overview}). The samples are generated every 15 minutes, resulting in 96 samples every day. We use one month of data for training (2880 samples) and evaluate on data of the next two months (5760 samples). The training data is divided in train-validation-test splits with 60-20-20 ratio.

\vspace{-.25cm}
\subsection{Training setup}

The model parameters are summarized in Table \ref{hyperparameters}. All models are implemented in Pytorch using Adam optimizer. For the ChebNet baseline \cite{hajgato2021pressure}, we set the learning rate of 3e-4 and weight decay of 6e-6. For our m-GCN models, we use learning rate of 1e-5 and no weight decay. We now describe the training setup of the ChebNet baseline and our m-GCN model for the two experiments, respectively. 

For the first experiment the models are trained for 2000 epochs. We set an early stopping criteria such that it stops after 250 epochs if the change in loss is no larger than 1e-6. We configure ChebNet similar to \cite{hajgato2021pressure}. Input is masked as per the sensor ratio and the mask is concatenated with the pressure values. Hence there are two node features. ChebNet can only use scalar edge fetaures, i.e. edge weights. Out of the three types of edge weights used by \cite{hajgato2021pressure} (binary, weighted, logarithmically weighted), we use the binary weights since other types did not increase performance. For our model (m-GCN), we did not perform an extensive hyperparameter search since we achieved considerably better results than ChebNet model of \cite{hajgato2021pressure} with a set of intuitive hyperparameter values. We use single hop configuration for Anytown and multi-hop architectures for C-Town and Richmond WDS. We only use masked pressure values as input i.e. one node feature. Further, we use two edge features namely pipe length and diameter. 

For the second in-depth evaluation on L-Town, we dropped the second node feature for ChebNet since this significantly improved the results. We use the ChebNet model configuration used for Richmond WDS by the authors. We train our m-GCN model with two configurations; one with the default single hop and the second with multiple hops as listed in Table \ref{hyperparameters}. For both m-GCN models, we add a third edge feature namely pressure reducing valves (PRVs) mask. PRVs are used at certain connections in a WDS to reduce pressure, hence these edges should be modeled differently. We use a binary mask to pass this information to the model that helps in improving the pressure estimation at neighboring nodes. We train all three models for 5000 epochs without early stopping.

\section{Results} \label{Results}

\begin{table}[t]
    \caption{Mean errors across nodes and samples across 5 different sensor configurations for 3 different ratios of sensors.}\label{results_baseline}
    \centering
    \resizebox{\textwidth}{!}{
    \begin{tabular}{|p{0.85cm}|p{2.05cm}|E|E|F|E|E|F|E|E|F|}
    \hline
    \multicolumn{2}{|c|}{WDS} & \multicolumn{3}{|c|}{Anytown} & \multicolumn{3}{|c|}{C-Town} & \multicolumn{3}{|c|}{Richmond} \\
    \hline
    Ratio & Error ($\times 10^{-3}$) & ChebNet & m-GCN & Diff & ChebNet & m-GCN & Diff & ChebNet & m-GCN & Diff \\
    \hline
    \multirow{3}{*}{0.05} & All & 54.19 & 53.15 & -1.04 & 12.88 & 9.77 & -3.11 & 4.34 & 2.17 & -2.17 \\
    & Sensor & 7.06 & 3.77 & -3.28 & 7.50 & 4.61 & -2.89 & 3.47 & 1.81 & -1.66 \\
    & Non-sensor & 56.44 & 55.50 & -0.94 & 13.16 & 10.04 & -3.12 & 4.38 & 2.19 & -2.19 \\
    \hline
    \multirow{3}{*}{0.1} & All & 35.43 & 34.85 & -0.57 & 8.16 & 5.47 & -2.69 & 3.86 & 1.93 & -1.93 \\
    & Sensor & 6.66 & 7.19 & 0.53 & 7.10 & 4.83 & -2.27 & 3.45 & 2.02 & -1.43 \\
    & Non-sensor & 38.3 & 37.62 & -0.68 & 8.28 & 5.55 & -2.73 & 3.90 & 1.92 & -1.98 \\
    \hline
    \multirow{3}{*}{0.2} & All & 14.98 & 13.51 & -1.47 & 7.05 & 5.58 & -1.47 & 3.24 & 1.59 & -1.65 \\
    & Sensor & 5.40 & 3.06 & -2.34 & 6.46 & 5.46 & -1.00 & 3.03 & 1.62 & -1.40 \\
    & Non-sensor & 17.11 & 15.83 & -1.28 & 7.20 & 5.61 & -1.59 & 3.29 & 1.59 & -1.71 \\
    \hline
    \end{tabular}
    }
    \vspace{-.5cm}
\end{table}

\paragraph{Comparison with spectral GCN-based approach}
First, we compare our model with the work of \cite{hajgato2021pressure} using their datasets and training settings. The results of the experiments on Anytown, C-Town and Richmond WDS are shown in Table \ref{results_baseline}. Here, we evaluate on the basis of mean relative absolute error given by:
\begin{equation}\label{eq_error}
    Error = \frac{1}{S\cdot N} \sum_{i=1}^{S} \sum_{v=1}^{N} \frac{| y_{iv} - \hat{y}_{iv} |}{y_{iv}}
\end{equation}
Since Anytown is a much smaller WDS, sensor ratios translate to very few sensors (0.05: 1 sensor, 0.1: 2 sensors, 0.2: 4 sensors). Hence, both models do not accurately estimate the pressures in these cases. The number of available sensors is comparatively bigger for both C-Town and Richmond WDS, even for the smallest ratio, thus naturally increasing performance. As can be seen, m-GCN outperforms ChebNet  \cite{hajgato2021pressure} by a considerable margin.

\begin{figure}[t]
    \includegraphics[width=\textwidth, keepaspectratio]{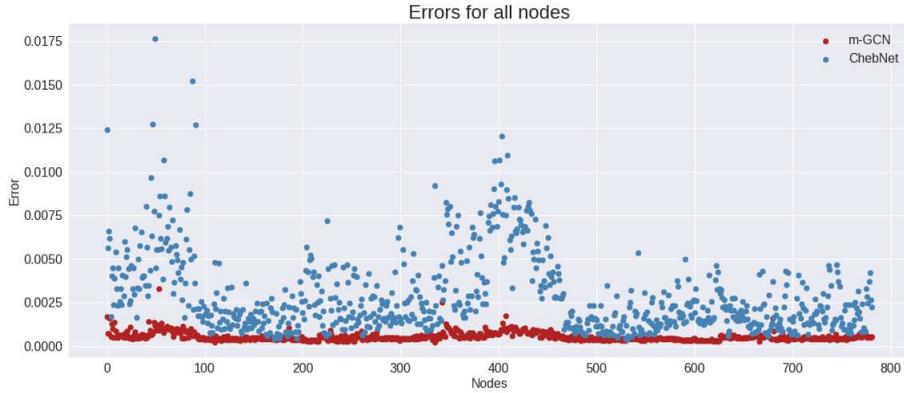}
    \vspace{-.832cm}
    \caption{Mean relative absolute errors for all nodes on noisy data for L-Town WDS.} \label{plot_error_noisy}
\end{figure}

\begin{figure}[!ht]
    \vspace{-.5cm}
    \includegraphics[width=\textwidth, keepaspectratio]{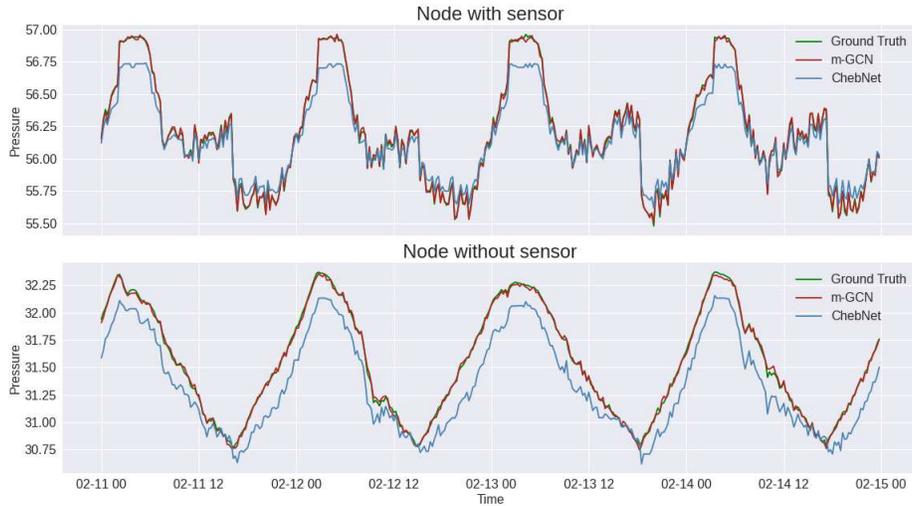}
    \vspace{-.832cm}
    \caption{Estimation results of m-GCN and ChebNet compared to ground truth on L-Town.} \label{pressure_pred_gcn_cheb}
    \vspace{-.5cm}
\end{figure}

\begin{figure}[t]
    \includegraphics[width=\textwidth, keepaspectratio]{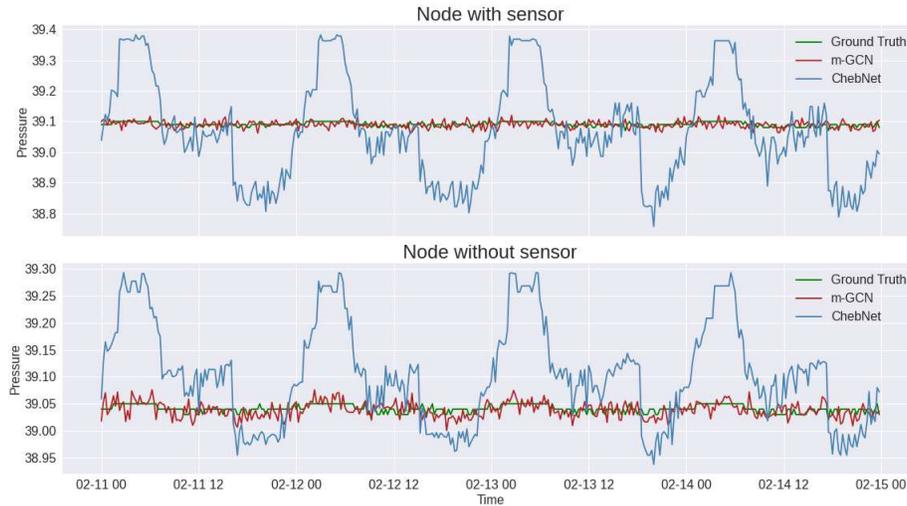}
    \vspace{-.832cm}
    \caption{Estimation results of m-GCN and ChebNet compared to ground truth on nodes from an area in L-Town with essentially stagnant pressure values.} \label{pressure_pred_gcn_cheb_erratic}
    \vspace{-.5cm}
\end{figure}

\vspace{-.25cm}
\paragraph{Detailed analysis on L-Town}
We present more in-depth analysis for the evaluation results on L-Town. Mean relative absolute errors for ChebNet and single-hop m-GCN models are plotted in Fig. \ref{plot_error_noisy}. Both models are trained on smooth data and evaluated on noisy realistic data. As can be seen, error values for m-GCN are much lower across all nodes compared to ChebNet. We plot time series of 4 days for a couple of nodes in Fig. \ref{pressure_pred_gcn_cheb}. The first node (top plot) has an installed sensor, hence the model gets the ground truth value as input and it has to only reconstruct it. The second node (bottom plot) does not have an installed sensor and the model gets zero-input. As depicted, m-GCN is able to successfully reconstruct and estimate both nodes. The results from ChebNet suffer considerable errors. There are areas in the L-Town WDS, where water levels are essentially stagnant with some noise. As shown in Fig. \ref{pressure_pred_gcn_cheb_erratic} our m-GCN is able to model those nodes correctly. In contrast, spectral convolutions do not take into account the graph structure and thus end up imposing the seasonality of nodes from other areas of the graph to the nodes in this area.  

Similar to our first experiment, we present mean relative absolute error values for all, sensor and non-sensor nodes for L-Town in Table \ref{results_ltown}. Our model produces significantly better results compared to the ChebNet. Since our model is based on neighborhood aggregation, the number of GCN layers required will continue to increase with the increasing size of the graphs. In order to reduce the number of layers and model parameters, we trained our model with only 10 GCN layers with 5 hops each. As evident, we are able to reduce the parameters by almost five times at the expense of some performance. Nevertheless, it is still significantly better than the baseline ChebNet model. Our main motivation for this is that the multi-hop approach makes the model more scalable to larger graphs. Further, it is a step towards developing a generalized version of the model that can work for different sensor configurations and/or different graph sizes without hyperparameter tuning and re-training.

\begin{table}
    \caption{Mean errors across nodes and samples on L-Town.}\label{results_ltown}
    \centering
    \resizebox{\textwidth}{!}{
    \begin{tabular}{|p{5cm}|G|G|G|}
    \hline
    \multirow{2}{*}{Model} & \multicolumn{3}{c|}{Error ($\times 10^{-3}$)} \\
    \cline{2-4}
     &  All & Sensor & Non-sensor \\
    \hline
     & \multicolumn{3}{c|}{Smooth Data} \\
    \hline
    ChebNet & 2.55 $\pm$ 2.87 & 2.38 $\pm$ 3.55 & 2.55 $\pm$ 2.83 \\
    m-GCN (45 x 1) & \textbf{0.39} $\pm$ 0.37 & \textbf{0.43} $\pm$ 0.52 & \textbf{0.39} $\pm$ 0.36  \\
    m-GCN (10 x 5) & 0.83 $\pm$ 0.68 & 0.74 $\pm$ 0.59 & 0.83 $\pm$ 0.69 \\
    \hline
    
     & \multicolumn{3}{c|}{Noisy Data} \\
    \hline
    ChebNet & 2.92 $\pm$ 3.35 & 2.78 $\pm$ 4.02 & 2.93 $\pm$ 3.32 \\
    m-GCN (45 x 1) & \textbf{0.54} $\pm$ 0.75 & \textbf{0.64} $\pm$ 1.06 & \textbf{0.53} $\pm$ 0.73 \\
    m-GCN (10 x 5) & 0.90 $\pm$ 0.82 & 0.81 $\pm$ 0.74 & 0.90 $\pm$ 0.83 \\
    
    \hline
    \end{tabular}
    }
    \vspace{-.5cm}
\end{table}

\section{Conclusion}

We have proposed a spatial GCN which is particularly suited for graph tasks on graphs with small node degree and sparse node features, since it is able to model long-term dependencies. We have demonstrated its suitability for node pressure inference based on sparse measurement values as an important and representative task from the domain of WDS, displaying its behavior for a number of benchmarks. Notably, the model generalizes not only across time windows, but also from noise-less toy demand signals to realistic ones. In addition to a very good performance overall, we also proposed first steps to target the challenge of scalability to larger graphs by introducing multi-hop architectures with considerably fewer parameters as compared to fully connected deep ones. In
future work, we will investigate the behavior for larger networks based on these first results. Moreover, unlike simulation tools in the domain, the GNN has the potential to generalize over different graphs structures including partially faulty ones. We will evaluate this capability in future work.

\vspace{-.25cm}
 \subsubsection{Acknowledgements} 
 We gratefully acknowledge funding from the European Research Council (ERC) under the ERC Synergy Grant Water-Futures (Grant agreement No. 951424). This research was also supported by the research training group “Dataninja” (Trustworthy AI for Seamless Problem Solving: Next Generation Intelligence Joins Robust Data Analysis) funded by the German federal state of North Rhine-Westphalia, and by funding from the VW-Foundation for the project \textit{IMPACT} funded in the frame of the funding line \textit{AI and its Implications for Future Society}.

%
%
%
\bibliographystyle{splncs04}
\vspace{-.25cm}
\bibliography{references}
\vspace{-.25cm}

\end{document}